\documentclass{bmvc2k}

\title{Deep Image Matting \\with Flexible Guidance Input}

\addauthor{Hang Cheng}{ch20721335@shu.edu.cn}{0}
\addauthor{Shugong Xu*}{shugong@shu.edu.cn}{0}
\addauthor{Xiufeng Jiang}{XiufengJiang@shu.edu.cn}{0}
\addauthor{Rongrong Wang}{wangrongrong@shu.edu.cn}{0}

\addinstitution{
 Shanghai Institute for Advanced Communication and Data Science(SICS),\\
 Shanghai University,\\
 Shanghai,\\
 200444,\\
 China\\
}

\runninghead{CHENG ET AL.}{Deep Image Matting with Flexible Guidance Input}

\def\etal{\emph{et al}\bmvaOneDot}

\usepackage{graphicx}
\usepackage{arydshln}
\usepackage{multirow}
\usepackage{soul}
\soulregister\cite7 
\soulregister\citep7 
\soulregister\citet7 
\soulregister\ref7 
\soulregister\pageref7 
\usepackage{url}

\begin{document}

\maketitle

\begin{abstract}
Image matting is an important computer vision problem. Many existing matting methods require a hand-made trimap to provide auxiliary information, which is very expensive and limits the real world usage. Recently, some trimap-free methods have been proposed, which completely get rid of any user input. However, their performance lag far behind trimap-based methods due to the lack of  guidance information. In this paper, we propose a matting method that use {\bf Flexible Guidance Input} as user hint, which means our method can use trimap, scribblemap or clickmap as guidance information or even work without any guidance input. To achieve this, we propose {\bf Progressive Trimap Deformation(PTD)} scheme that gradually shrink the area of the foreground and background of the trimap with the training step increases and finally become a scribblemap. To make our network robust to any user scribble and click, we randomly sample points on foreground and background and perform curve fitting. Moreover, we propose {\bf Semantic Fusion Module(SFM)} which utilize the Feature Pyramid Enhancement Module(FPEM) and Joint Pyramid Upsampling(JPU) in matting task for the first time. The experiments show that our method can achieve state-of-the-art results comparing with existing trimap-based and trimap-free methods. Our demo is at \url{https://github.com/Charch-630/FGI-Matting}.

\end{abstract}


\section{Introduction}
\label{sec:intro}

Image Matting is an important computer vision problem which aims to precisely predict an alpha matte to separate the foreground object from the background. The problem has already been studied by both academia and industry for years and has many applications in image processing and film production. Ordinarily, the Image Matting task is modeled to solve the following equation known as the Composition Equation.


\begin{small}
\begin{equation}
I_{i}=\alpha _{i}F _{i} + \left( 1 -  \alpha _{i}\left) B_{i},\right. \right.\ \ \ \ \ \ \alpha\in [0, 1]
\end{equation}
\end{small}

 \vspace{-0.7cm}
where $I$ denotes the input Image, $\alpha$ refers to the alpha matte that shows the opacity of the foreground object, $F$ and $B$ represent the foreground and background respectively, $i$ means per pixel location. As only the input RGB image is given and the algorithm has to solve the rest 7 values for each pixel, the problem is highly ill-posed.

In order to address this problem, most previous methods also require a hand-made trimap to provide auxiliary information. The trimap uses white($\alpha=1$), gray($\alpha=0.5$) and black($\alpha=0$) to define the foreground, transition area and background respectively. In the past few years, many methods~\cite{levin2007closed,zheng2009learning,chen2013knn,xu2017deep,lu2019indices,cai2019disentangled,hou2019context,li2020natural} using trimap as input have achieved very good accuracy and very few trimap-free methods can surpass them. However, when it comes to application, drawing a suitable and correct trimap requires some skills and much time, which is hassle for users who don't have any prior knowledge about matting.

In the past few years, some trimap-free methods have been proposed. These methods~\cite{zhang2019late,qiao2020attention} hope to capture both semantic feature and texture details from the RGB input by end-to-end training on large-scale dataset. However, their performance still lag far behind trimap-based methods and these methods are controversial. The main reason is that without guidance input, the network is confused about which part of the input image is foreground area and thus affect subsequent detail feature extraction. 

\begin{figure}
\includegraphics[width=\textwidth]{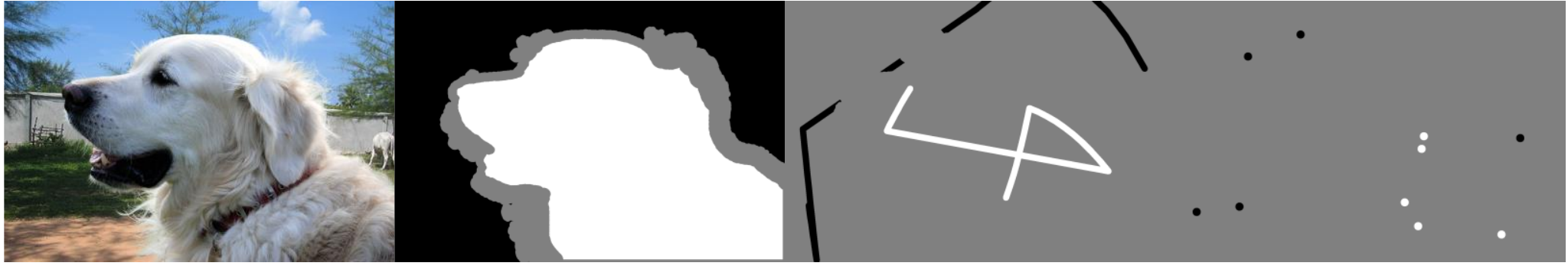}
\caption{Different guidance input. Trimap, scribblemap and clickmap.}
\label{fig1}
\end{figure}

To solve this, we propose a matting method that use flexible guidance input as user hint, which means our method works not only for trimap input, but also for scribble and click input and even no guidance input(see Fig.~\ref{fig1}). To achieve this, we introduce a data augmentation scheme called Progressive Trimap Deformation(PTD) that gradually shrink the area of the foreground and background of the trimap with the training step increases, and the shape of the foreground and background will eventually become scribbles. Moreover, to make our network robust to any user scribble and click, we propose to randomly sample points on foreground and background and perform curve fitting to simulate human input scribbles. As a result, experiments show that our method can achieve state-of-the-art results compared with previous trimap-based and trimap-free methods. Compared to previous trimap-based matting methods, ours reduces the complexity of the guidance input while ensuring the accuracy. When applied to real-world scene, for foreground objects of different difficulty, users can flexibly choose guidance input of different complexity levels. For example, for hard foreground objects, we can draw a trimap to give the network more details. For those of moderate difficulty, we can draw scribbles to save some time. For simple and salient objects, we just need a few clicks or even no guidance is needed. More importantly, our method can easily be applied to train other trimap-based methods, making them only require scribble or click input.

Moreover, in matting tasks, advanced semantics from deep levels of the backbone indicate foreground category and profiles, while low-level features contain texture and detail information. High-level semantic features can guide low-level features to correctly predict the details of the foreground region. In order to extract advanced semantics efficiently, we propose Semantic Fusion Module(SFM) inspired by FPEM~\cite{wang2019efficient} and JPU~\cite{wu2019fastfcn}. The SFM module can enhance semantic features by extracting multi-scale information from the backbone, then the advanced semantics of different scales can be joint upsampled and fused. 

To sum up, our contribution mainly includes the following points:
\begin{itemize}
    \vspace{-0.2cm}
    \item[$\bullet$]We propose a matting method based on Flexible Guidance Input, which means our method can use trimap, scribblemap or clickmap as guidance information. Moreover, Our method can even work without any guidance input. The experiments verify that our method can achieve state-of-the-art results comparing with existing trimap-based and trimap-free methods.
    \vspace{-0.2cm}
    \item[$\bullet$]We propose a data augmentation scheme called Progressive Trimap Deformation(PTD) that gradually shrink the area of the foreground and background of the trimap during training and eventually become a scribblemap. To make our network robust to any user scribble and click, we randomly sample points on foreground and background and perform curve fitting to mimic human input scribbles.
    \vspace{-0.2cm}
    \item[$\bullet$]In order to extract advanced semantics efficiently, we propose Semantic Fusion Module(SFM) which utilize the FPEM~\cite{wang2019efficient} and JPU~\cite{wu2019fastfcn} modules in matting tasks for the first time. The SFM module extract multi-scale information from the backbone and then fuse and joint upsample them to enhance the advanced semantics.
\end{itemize}

\vspace{-0.5cm}
\section{Related Work}
\vspace{-0.2cm}
\subsection{Trimap-based Matting methods.}Most existing matting methods require a trimap as an auxiliary input. Traditional matting methods can be categorized into two types: sampling based and propagation based. Sampling-based methods~\cite{chuang2001bayesian,feng2016cluster,gastal2010shared} first model foreground and background statistics through sampling pixels in the given foreground and background area, then solve the composition equation to get alpha matte. Propagation based methods~\cite{chen2013knn,levin2007closed} propose to propagate the alpha value from the given foreground and background region to unknown area. In the past few years, deep learning based methods have been proved successful in solving image matting problems. Xu \etal~\cite{xu2017deep} proposed an encoder-decoder structure with RGB image and a trimap as input to predict alpha matte, and created a matting dataset with various foregrounds composited to background images. Hou \etal~\cite{hou2019context} propose to use two decoder to predict foreground color and alpha simultaneously. Li \etal~\cite{li2020natural} propose a u-net structure with a guided contextual attention block and they achieved better results.

\vspace{-0.2cm}
\subsection{Trimap-free Matting methods.}Recently, some trimap-free matting approaches have emerged. Some of them propose to train on large-scale dataset to completely get rid of trimap. Zhang \etal~\cite{zhang2019late} use two decoder branches to predict foreground and background classification respectively and then fuse them together. Its input is only an RGB image. Qiao \etal~\cite{qiao2020attention} propose to use spatial and channel attention to filter high-level semantics and appearance feature. Others find easier form of guidance input instead of trimap or use semantic information. Liu \etal~\cite{liu2020boosting} propose to use a course mask as guidance input. Chen \etal~\cite{chen2018semantic}propose to automatic generate trimap using semantic information. Sengupta \etal~\cite{sengupta2020background} and Lin \etal~\cite{lin2020real} propose to take another photo of the background as auxiliary input. Wei \etal~\cite{wei2020improved} propose to use user clicks as foreground and background hints.

\section{Methodology}In this Section, we explain the details of our method. We first elaborate our Progressive Trimap Deformation scheme, and then we introduce our Semantic Fusion Module.

\begin{figure}
\setlength{\abovecaptionskip}{-0.5cm}
\setlength{\belowcaptionskip}{-0.4cm}
\includegraphics[width=\textwidth]{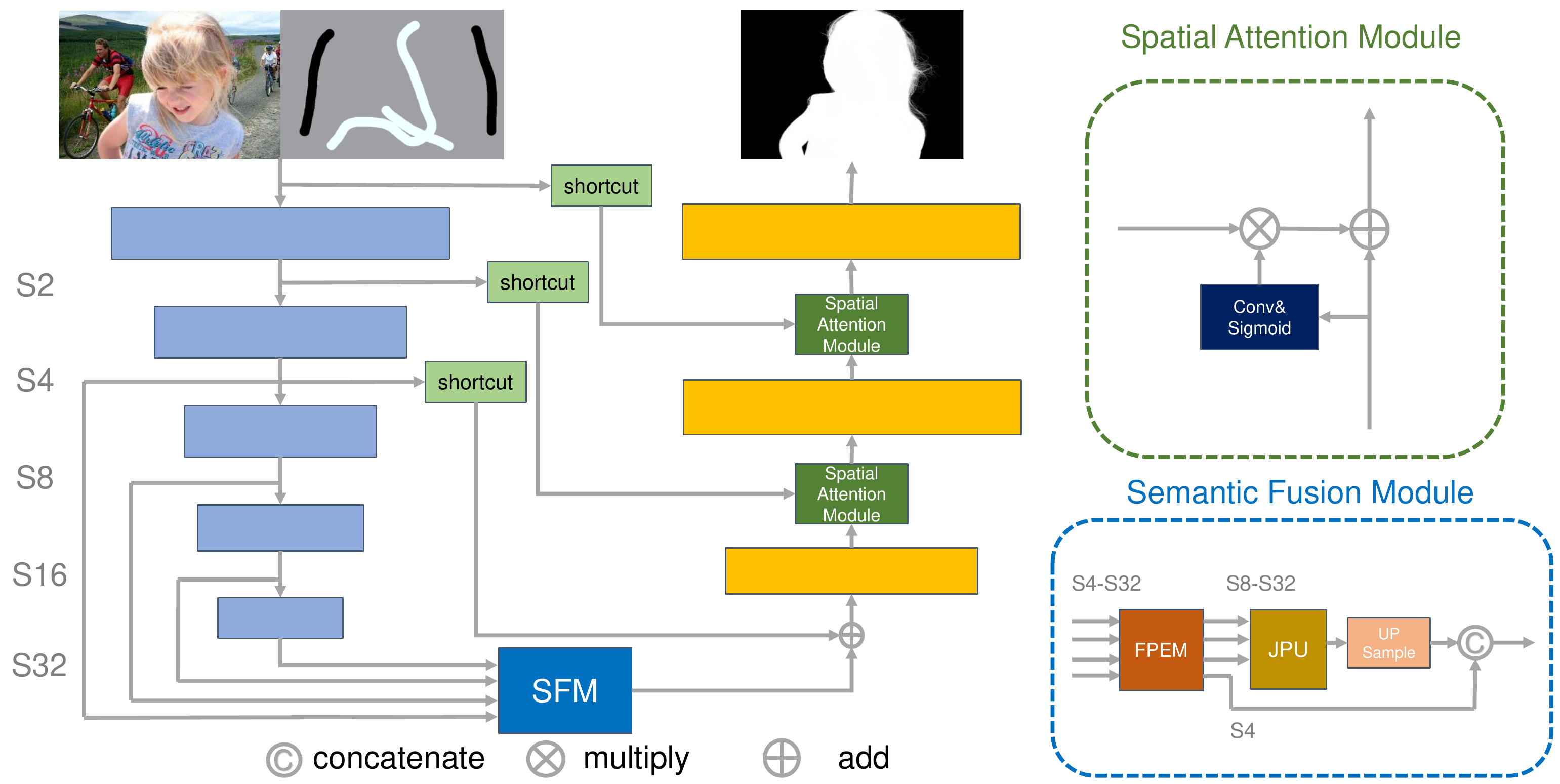}
\caption{Our proposed network structure. Our network is based on a U-Net architecture~\cite{li2020natural}.}
\label{fig2}
\end{figure}

\vspace{-0.4cm}
\subsection{Progressive Trimap Deformation}
To find a simpler form of guidance input than trimap, Wei \etal~\cite{wei2020improved} use clickmap to train the network. The clickmap use white or black circles of radius r to indicate foreground or background hint respectively. Intuitively, the definition of the clickmap is similar to trimap. Both of them divide the image into absolute foreground, absolute background and transition area. The difference is that clickmap provides far less guidance information than trimap. Wei \etal~\cite{wei2020improved} directly use clickmap to train the network and the result is better than trimap-free methods but worse than trimap-based methods. Based on the above analysis, using clickmap is much easier than trimap but the result is less accurate. This motivates us to train a network that works for all kinds of guidance input between the trimap and clickmap(see Fig.~\ref{fig1}). We leave the trade-off between accuracy and difficulty to user, enabling flexible guidance input. The more accurate the guidance input is, the better result it can achieve. To do this, we gradually shrink the foreground and background area of the trimap with the training step increases. In this way, the network can learn to leverage guidance information rather than being restricted to the domains of trimap or clickmap. During the shrinking process, we slowly reduce FG and BG area to make the network better adapt to less guidance information, and this gentle process can make the network converge better. In our method, we shrink the trimap to scribblemap during training. Scribblemaps are more varied in shape compared with clickmaps, which can bring more challenge to the network.


\vspace{-0.2cm}
\subsubsection{Network Architecture}Our network is based on a U-Net architecture~\cite{li2020natural} (see Fig.~\ref{fig2}) and we use ResNet34 as our backbone. The input RGB image is concatenated with a one-channel guidance map, which can be a trimap, scribblemap or clickmap. In the encoder part, the stride-1, stride-2 and stride-4 output of the encoder are fed into shortcut blocks. The shortcut blocks consist of two stride-1 convolution layers followed by batch normalization and Relu. These shortcut blocks are used to process low-level texture features. To leverage multi-scale information from the backbone, stride-4, stride-8, stride-16 and stride-32 output of the encoder are fed to our proposed SFM module to enhance high-level semantic features. In the decoder part, we use dilated convolutions to upsample the feature map. Since the high-level semantic feature indicate foreground category and profiles, we can use it to filtrate redundant low-level texture feature. Thus we use spatial attention module to process low-level features from the shortcut blocks.

\begin{figure}
\setlength{\abovecaptionskip}{-0.5cm}
\setlength{\belowcaptionskip}{-0.7cm}
\includegraphics[width=\textwidth]{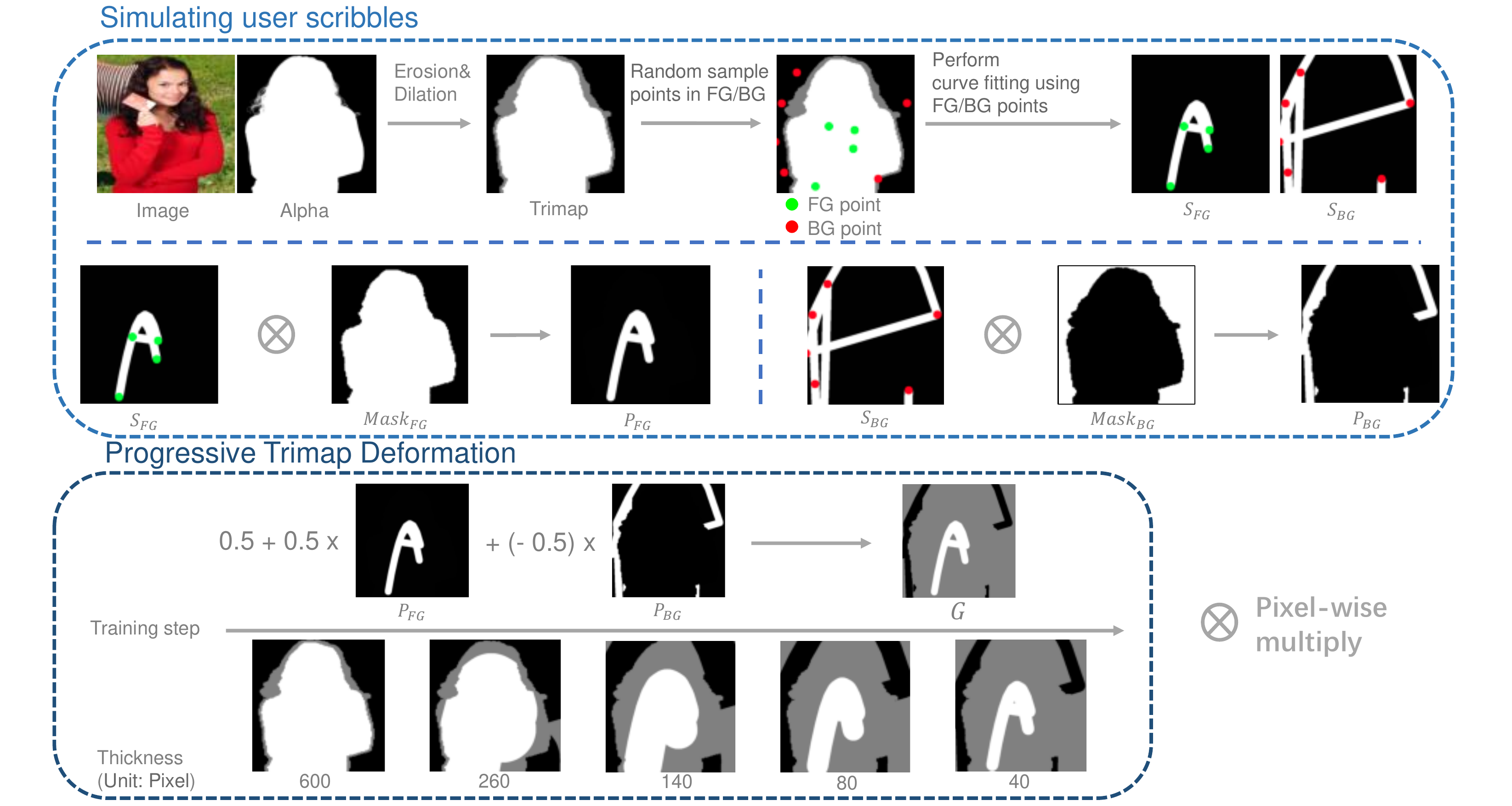}
\caption{The process of our proposed Progressive Trimap Deformation.}
\label{fig3}
\end{figure}

\vspace{-0.4cm}
\subsubsection{Simulating User Scribbles.}As mentioned above, we propose to shrink the trimap to scribblemap during training. To implement this process, we first need to use trimap to generate the target scribblemap(see Fig.~\ref{fig3}). In each training step, the same as the previous trimap-based matting methods, we generate the trimap by using dilation and erosion operation on groundtruth alpha matte. Then we randomly select a total of up to 10 points on foreground and background area of each trimap. To avoid the points being too close to each other, we set a threshold of 50 pixels between each two points. After that, we use FG points and BG points to do curve fitting respectively. In detail, for FG points, we iteratively retrieve 3 points from the sampled FG points at a time. Then we use a cubic function to fit the curve through the 3 points. Finally we draw all fitted curves with a certain thickness on one graph to get $S_{FG}$. Using the same way, we can get $S_{BG}$. Now we can make sure that most part of the scribbles are in FG or BG area respectively, but we still have to deal with the excess part. So we simply use $Mask_{FG}$ and $Mask_{BG}$ from the trimap to restrict $S_{FG}$ and $S_{BG}$ using function $P_{FG} = S_{FG}*Mask_{FG}$ and $P_{BG} = S_{BG}*Mask_{BG}$. $P_{FG}$ and $P_{BG}$ are the final generated FG scribbles and BG scribbles.
 

\subsubsection{Foreground and Background Deformation.}After the foreground scribble mask and background scribble mask has been generated, we then generate the scribblemap using the function below.
\begin{small}
\begin{equation}
G = 0.5 + 0.5*P_{FG} + (-0.5)*P_{BG}
\end{equation}
\end{small}
To gradually shrink the trimap to scribblemap, we can simply change the thickness of the scribbles during training. At the beginning of training, the scribbles are thick enough to cover all FG and BG regions, which makes our guidance map almost the same as the trimap. As the training step increases, we gradually decrease the thickness of the scribbles. As we can see in Fig.~\ref{fig3}, the area of FG and BG are gradually reduced. In this way, we gradually give less guidance information to the network during training which enhance the network's ability to distinguish the foreground from the background.

\begin{figure}
\setlength{\abovecaptionskip}{-0.5cm}
\setlength{\belowcaptionskip}{-0.7cm}
\includegraphics[width=\textwidth]{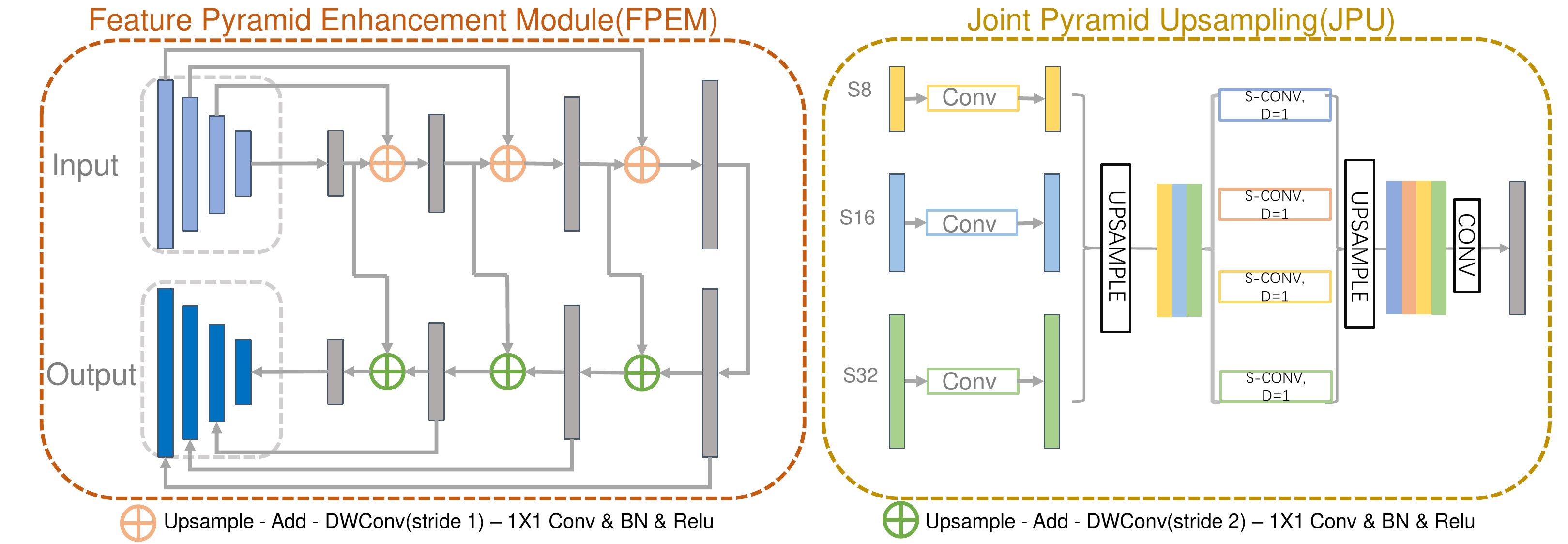}
\caption{The architecture of FPEM~\cite{wang2019efficient} and JPU~\cite{wu2019fastfcn} module.}  
\label{fig4}
\end{figure}

\subsubsection{Loss function.}Our loss function is based on ~\cite{wei2020improved, zhang2019late}. We use three types of loss functions.

\begin{small}
\begin{equation}
L(\hat{\alpha},\alpha) = \frac{1}{|K|}\sum_{i\in K}^{}(\hat{\alpha}_{i} -\alpha_{i})^{2} +\frac{1}{|T|}\sum_{i\in T}^{}|\hat{\alpha}_{i}-\alpha_{i}|+L_{grad}(\hat{\alpha},\alpha)
\end{equation}
\end{small}

Where $K$ denotes foreground and background region and $T$ denotes transition region. $\hat{\alpha}$ and $\alpha$ indicate the predicted alpha matte and ground-truth. We apply $\ell _{2}$ loss in foreground and background region and apply $\ell _{1}$ loss in transition region. The $\ell _{2}$ loss in foreground and background region improves the prediction of object contour and the $\ell _{1}$ loss in transition area helps the detail prediction. At the beginning of training, $\ell _{2}$ loss is more sensitive than $\ell _{1}$ loss, which makes the network focus on foreground and background areas. As loss begins to converge, $\ell _{1}$ loss will become more sensitive than $\ell _{2}$ loss, the network will focus on the details in transition area.

\begin{small}
\begin{equation}
L_{grad}(\hat{\alpha},\alpha) = \frac{1}{|I|}\sum_{i\in I}^{}|\bigtriangledown (\hat{\alpha}_{i})-\bigtriangledown(\alpha_{i})|
\end{equation}
\end{small}

$L_{grad}$ is defined as $\ell _{1}$ loss on the gradient of $\hat{\alpha}$ and $\alpha$. $I$ represents the whole image area. $L_{grad}$ makes the network produce sharper alpha mattes. We compute the gradient by convolving the alpha matte with first-order Gaussian derivative filter.

\subsection{Semantic Fusion Module}In matting tasks, advanced semantics from deep levels of the backbone indicate foreground category and profiles, which can guide low-level features to correctly predict the details of the foreground region. In order to extract advanced semantics efficiently, we propose to leverage multi-scale information from the backbone. Another problem is that in order to obtain a high-resolution output alpha matte, the decoder usually use dilated convolution to upsample the feature map, which brings heavy computation. To solve these two problems, we propose Semantic Fusion Module(SFM) which is composed of two modules, Feature Pyramid Enhancement Module(FPEM)~\cite{wang2019efficient} and Joint Pyramid Upsampling(JPU)~\cite{wu2019fastfcn}(see Fig.~\ref{fig4}). For the first time, we use the FPEM and JPU modules in matting tasks. The FPEM is a cascadable U-shaped module consists of up-scale enhancement and down scale enhancement using the feature pyramid. By fusing the low-level and high-level information, FPEM is able to enhance multi-scale features. In JPU module, four separable convolutions with different dilation rates are used to extract information from multi-level feature map. The JPU module is designed to obtain a feature map similar to the result of using dilated convolution but with less computation cost. In SFM module, we simply cascade FPEM and JPU together(see Fig.~\ref{fig2}). Thus we can first use FPEM to extract multi-scale information, then use JPU to upsample the feature maps with less computational complexity and better performance.

\begin{table}
\footnotesize
\centering
\caption{Results on Composition-1k test set. We use dashlines to divide these methods into three categories, from top to button: trimap-based, trimap-free and ours.}\label{tab1}
\begin{tabular}{l|llll}
\hline
Methods & SAD$\ \ \ \ \ \ \ \ \ $   & MSE$\ \ \ \ \ \ \ \ \ $   & Grad$\ \ \ \ \ \ \ \ \ $  & Conn$\ \ \ \ \ \ \ \ \ $  \\ \hline
KNN Matting\cite{chen2013knn}            & 175.4 & 0.010 & 124.1 & 176.4 \\
ClosedForm\cite{levin2007closed}             & 168.1 & 0.091 & 126.9 & 167.9 \\
Alphagan\cite{lutz2018alphagan}            & 90.9 & 0.018 & 93.9 & 95.2 \\
DIM\cite{xu2017deep}                    & 48.8  & 0.008 & 31.0  & 50.3  \\
IndexNet Matting\cite{lu2019indices}       & 45.8  & 0.013 & 25.9  & 43.7  \\
AdaMatting\cite{cai2019disentangled}             & 41.7  & 0.010 & 16.9  & -     \\
Context-Aware Matting\cite{hou2019context} $\ \ \ \ \ \ \ \ \ \ \ \ \ \ \ \ \ \ $  & 35.8  & 0.082 & 17.3  & 33.2  \\
GCA Matting\cite{li2020natural}            & 35.3  & 0.009 & 16.9  & 32.5  \\
GCA Matting(Scribblemap\_test)            & 48.7  & 0.025 & 22.9  & 39.5  \\
GCA Matting(Clickmap\_test)            & 53.2  & 0.029 & 24.7  & 41.4  \\\hdashline
Late Fusion\cite{zhang2019late}            & 58.3  & 0.011 & 41.6  & 59.7  \\
HAttMatting\cite{qiao2020attention}            & 44.0  & 0.007 & 29.2  & 46.4  \\ \hdashline
Ours(Trimap\_test)           & 30.19  & 0.0061 & 13.07  & 26.66  \\
Ours(Scribblemap\_test)      & 32.86  & 0.0090 & 14.18  & 29.09  \\
Ours(Clickmap\_test)         & 34.67  & 0.0112 & 15.45  & 30.96  \\
Ours(No\_guidance\_test)      & 36.36  & 0.0141 & 15.23  & 32.76  \\ \hline
\end{tabular}
\end{table}

\vspace{-0.4cm}
\section{Experiments}In this section, we report the test results of our method. We compare our method with existing matting methods on DIM dataset~\cite{xu2017deep} and conduct an ablation study of our method.

\begin{figure}
\includegraphics[width=\textwidth]{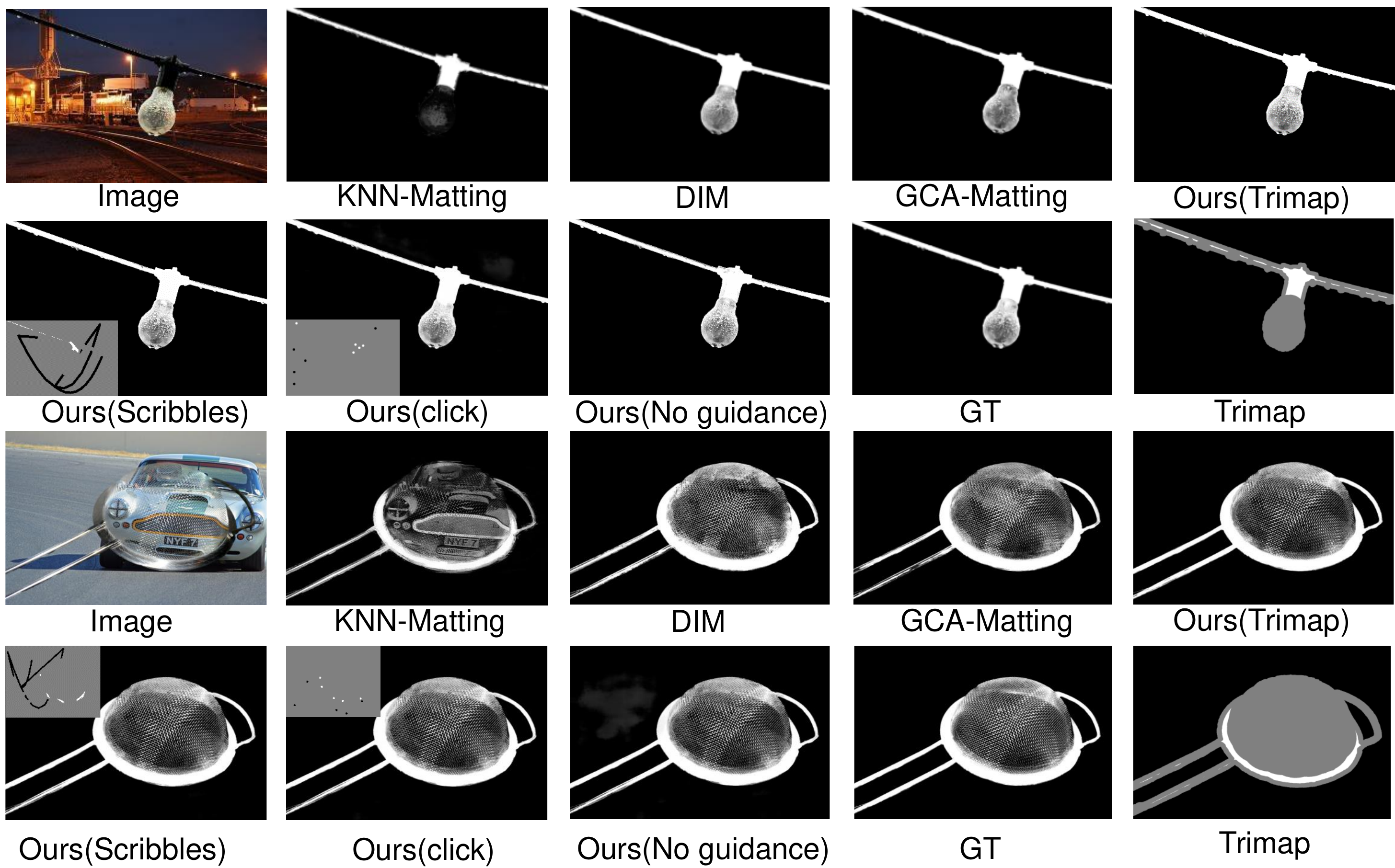}
\caption{Visual comparisons on Composition-1k} 
\label{fig5}
\end{figure}

\vspace{-0.2cm}
\subsection{Implementation Details} We simply follow GCA-matting\cite{li2020natural}'s training strategy and apply their data augmentation methods during training. The data augmentation operations include random affine transformation, random cropping, random jitters and random resize and our PTD scheme. The input resolution of the network is $512*512$. During training, we set the initial thickness of the curves to be 800 and gradually decrease to 40 before the step of 530k. The relationship between the thickness and step is an exponential function which makes the thickness decreases slower at the later stage of training to avoid performance drop on trimap. After that, we then train the network with thickness of 40 for 70k steps. The base learning rate is set to $5*10^{-4}$ with cosine learning rate scheduler. We set the $\beta_{1}$ and $\beta_{2}$ of the Adam optimizer to 0.5 and 0.999. We use batch size of 10 in total on 2 GPUs.

\subsection{Experiments on DIM Dataset}
The DIM dataset\cite{xu2017deep} contains 43100 synthetic image with 431 unique foreground objects. The Composition-1k test set consists of 1000 synthetic images. In addition to testing with trimap, we also use the trimap in Composition-1k to generate scribblemap test set and clickmap test set in the same way as in our PTD. To obtain clickmaps, we simply draw circles of diameter 40 with 1 and 0 on foreground and background sampled points respectively. We also test our network without guidance information by using one-channel tensor with a value of 0.5 on each pixel as guidance input.

We follow the previous methods to evaluate the results by using Sum of Absolute Differences(SAD), Mean Squared Error(MSE), Gradient(Grad) and Connectivity(Conn) metrics, and all metrics are to be minimized. As shown in Table.~\ref{tab1}, compared with previous trimap-based methods, our method outperforms all previous methods when tested with trimap. When using clickmap as guidance information, our method is still superior to GCA-Matting in the SAD metric. Moreover, the previous trimap-based methods all need an accurate trimap, while ours can get a better result with just a few clicks or scribbles. We also show the results of GCA-matting tested on our generated scribblemap test set and clickmap test set in Table.~\ref{tab1}. When both tested on scribblemap or clickmap, our method can outperform GCA-Matting on all metrics. This is largely due to our PTD scheme, while GCA-matting is trained only on trimap. Compared with previous trimap-free methods, ours(No Guidance test) surpasses previous state-of-the-art methods by a large margin. The reason is mainly because previous trimap-free methods didn't use any guidance information in the training stage while ours gradually reduce guidance information.

\begin{table}
\setlength{\abovecaptionskip}{-0.1cm}
\scriptsize
\centering
\caption{Our method tested using different Scribblemap test sets and Clickmap test sets.}\label{tab2}
\begin{tabular}{l|llll|lllll}
\hline
\multicolumn{5}{c|}{Scribblemap}     & \multicolumn{5}{c}{Clickmap}                              \\ \hline
Num & SAD   & MSE    & Grad  & Conn  & \multicolumn{1}{l|}{Num} & SAD   & MSE    & Grad  & Conn  \\ \hline
1   & 32.86 & 0.0090 & 14.18 & 29.09 & \multicolumn{1}{l|}{1}   & 34.67 & 0.0112 & 15.45 & 30.96 \\
2   & 33.08 & 0.0093 & 14.22 & 29.32 & \multicolumn{1}{l|}{2}   & 34.45 & 0.0109 & 15.24 & 30.74 \\
3   & 32.70 & 0.0086 & 14.10 & 28.91 & \multicolumn{1}{l|}{3}   & 34.46 & 0.0110 & 15.27 & 30.74 \\ \hline
$\sigma$ & 0.1557 & 0.00029 & 0.0498 & 0.1678 & \multicolumn{1}{l|}{$\sigma$} & 0.1014 & 0.00012 & 0.0927 & 0.1037 \\ \hline
\end{tabular}
\end{table}

\begin{table}
\scriptsize
\centering
\caption{Ablation study of our method. Baseline: ResNet34 U-net with FPN structure.}\label{tab3}
\begin{tabular}{l|l|llll}
\hline
Num & Method                                      & SAD$\ \ $   & MSE$\ $     & Grad$\ \ $  & Conn$\ $  \\ \hline
1      & Baseline + Trimap\_train + Trimap\_test                       &  31.87 & 0.0068  & 13.48  &  28.55   \\\hdashline
2      & Baseline + FPEM + Trimap\_train + Trimap\_test                 &  32.94 & 0.0069  & 14.42  &  29.85   \\
3      & Baseline + JPU + Trimap\_train + Trimap\_test                  &  33.48 & 0.0072  & 14.89  &  30.49 \\\hdashline
4      & Baseline + SFM + Trimap\_train + Trimap\_test                   & 30.99 & 0.0064  & 12.81 & 27.54 \\
5      & Baseline + SFM + Trimap\_train + Scribblemap\_test                   & 44.33 & 0.0203  & 16.40 & 39.12 \\
6      & Baseline + SFM + Trimap\_train + Clickmap\_test                   & 48.53 & 0.0250  &  17.84 & 42.92 \\
7      & Baseline + SFM + Trimap\_train + No\_guidance\_test & 52.97 &  0.0323  & 18.56 & 47.43 \\ \hdashline
8      & Baseline + PTD + Trimap\_test               &  32.86 &  0.0069  & 14.01 & 29.92 \\
9      & Baseline + PTD + Scribblemap\_test             & 35.96 &  0.0118  & 14.79 &  32.92 \\
10      & Baseline + PTD + Clickmap\_test             & 37.44 & 0.0138  & 15.41 & 34.50 \\
11      & Baseline + PTD + No\_guidance\_test             & 38.67 & 0.0165  & 15.34 & 35.91 \\\hdashline
12      & Baseline + SFM + PTD + Trimap\_test               &  30.19 &  0.0061  & 13.07 & 26.66 \\
13      & Baseline + SFM + PTD + Scribblemap\_test             & 32.86 &  0.0090  & 14.18 &  29.09 \\
14      & Baseline + SFM + PTD + Clickmap\_test             & 34.67 & 0.0112  & 15.45 & 30.96 \\
15      & Baseline + SFM + PTD + No\_guidance\_test             & 36.36 & 0.0141  & 15.23 & 32.76 \\\hline
\end{tabular}
\end{table}

\begin{figure}
\includegraphics[width=\textwidth]{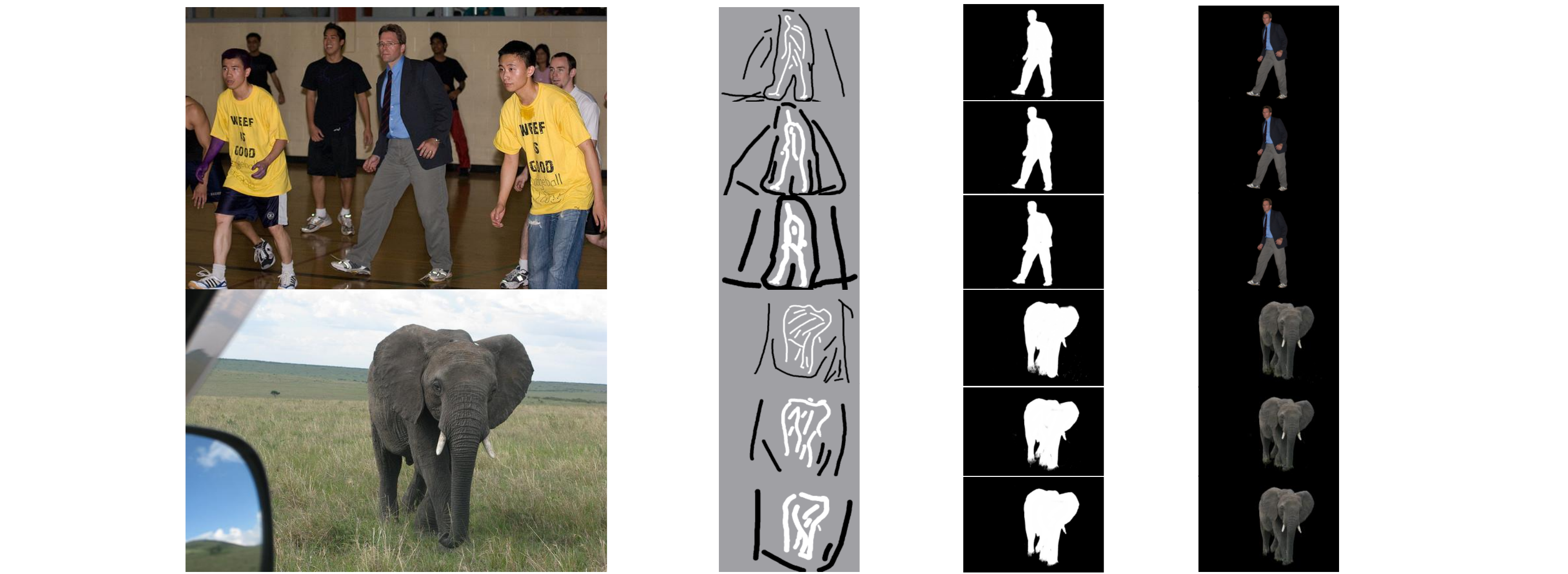}
\caption{Visual results of real images. Each column from left to right: input images, guidance map, predicted alpha matte, foreground object result.} 
\label{fig6}
\end{figure}

When comparing our method using different guidance input, we can find that the results get worse as the guidance information decreases. The experiments prove that our method can enhance the network's ability to make full use of the guidance information to distinguish the foreground from the background. The more accurate the guidance input is, the better result it can achieve. Even without any guidance information, our method still shows great robustness. We provide some comparison results in Fig.~\ref{fig5}. We also test our method by using real images, the results are shown in Fig.~\ref{fig6}.

We also test our method using different generated scribblemap test sets and clickmap test sets of Composition-1k. The results are in Table.~\ref{tab2}. The scribblemaps in different scribblemap test sets of the same image have the same thickness but different shapes, so do clickmaps. The standard deviation of all metrics in "scribblemap" column are very small, and "clickmap" column has the same phenomenon, proving that our method is robust to guidance maps of the same level(e.g. scribblemap, clickmap) but with different shape.

\subsection{Ablation Studies}To show the effectiveness of our SFM and PTD, we conduct ablation studies in Table.~\ref{tab3}. Compare 1,2,3 and 4, we can find that the SFM module can improve the network's performance on trimap, while using only FPEM or JPU will make the performance worse. Compare 8,9,10,11 and 12,13,14,15, we can also find that SFM can enhance the performance when training with PTD. Compare 4 to 12, the results show that when both tested on trimap, training with PTD is slightly better than training with trimap. This is probably because our PTD scheme can enhance the network's ability to leverage guidance information. Moreover, when testing with scribbles, clicks and no guidance information, our PTD scheme(13,14,15) can achieve far better results than training with trimap(5,6,7). The results show that our PTD scheme can ensure the performance on trimap, and can still achieve high accuracy with reduced guidance information. Note that we only shrink the trimap to scribblemap, but the performance on clickmap and no guidance is also enhanced. This is probably because during the shrinking process, the gray areas continue to expand, and the network is forced to leverage all given FG and BG information and improve the performance in gray areas. When tested with less guidance information, the network will still make full use of the given hints and try to predict the alpha matte in gray areas.



\begin{table}
\scriptsize
\centering
\caption{Results of our PTD applied on GCA and DIM matting. Note that the GCA and DIM in this table are based on our own training results.}\label{tab4}
\begin{tabular}{l|llll|llll}
\hline
                   & \multicolumn{4}{c|}{GCA}       & \multicolumn{4}{c}{DIM}        \\ \cline{2-9} 
                   & SAD   & MSE    & Grad  & Conn  & SAD   & MSE    & Grad  & Conn  \\ \hline
Trimap\_test       & 35.03 & 0.0086 & 16.54 & 30.78 & 56.77 & 0.0166 & 27.34 & 50.37 \\
Scribblemap\_test  & 46.92 & 0.0205 & 19.92 & 36.95 & 81.98 & 0.0523 & 33.59 & 65.90 \\
Clickmap\_test     & 52.57 & 0.0251 & 21.35 & 39.46 & 87.02 & 0.0596 & 35.17 & 68.83 \\
No\_guidance\_test & 60.15 & 0.0359 & 22.86 & 44.06 & 90.10 & 0.0654 & 35.52 & 71.32 \\ \hline
                   & \multicolumn{4}{c|}{GCA+PTD}   & \multicolumn{4}{c}{DIM+PTD}    \\ \cline{2-9} 
                   & SAD   & MSE    & Grad  & Conn  & SAD   & MSE    & Grad  & Conn  \\ \hline
Trimap\_test       & 32.71 & 0.0082 & 14.01 & 27.78 & 52.50 & 0.0150 & 29.42 & 46.11 \\
Scribblemap\_test  & 34.08 & 0.0111 & 13.93 & 28.65 & 61.80 & 0.0298 & 31.07 & 51.79 \\
Clickmap\_test     & 37.70 & 0.0172 & 14.90 & 30.98 & 64.12 & 0.0336 & 31.83 & 53.08 \\
No\_guidance\_test & 41.05 & 0.0242 & 15.63 & 32.57 & 65.67 & 0.0370 & 31.63 & 54.16 \\ \hline
\end{tabular}
\end{table}

We also apply our PTD scheme on other trimap-based methods to enhance their performance on multiple kinds of guidance inputs. In Table.~\ref{tab4}, we show the results of applying our PTD on GCA and DIM matting. Experiments show that our PTD can improve the performance of GCA and DIM on trimap, and significantly enhance the performance on scribblemap, clickmap and no extra guidance.

\section{Conclusion}In this paper, we propose a matting method that use flexible guidance input as user hint. Our method can use trimap, scribblemap or clickmap as guidance information or even work without any guidance input. To achieve this, we introduce our Progressive Trimap Deformation scheme and Semantic Fusion Module. The experiments show that our method can achieve state-of-the-art results compared with existing trimap-based and trimap-free methods. 

\ 

{\bf Acknowledgments.}  This work was supported by the National Natural Science Foundation of China, No.61871262.


\bibliography{egbib}
\end{document}